\documentclass[tablecaption=bottom,pmlr]{jmlr}

 \usepackage{rotating}%

\usepackage{booktabs}
\usepackage[load-configurations=version-1]{siunitx} %

\theorembodyfont{\upshape}
\theoremheaderfont{\scshape}
\theorempostheader{:}
\theoremsep{\newline}

\jmlrvolume{148}
\firstpageno{155}
\jmlryear{2021}
\jmlrsubmitted{submission date}
\jmlrpublished{publication date}
\jmlrworkshop{NeurIPS 2020 Preregistration Workshop} %

\usepackage[utf8]{inputenc} %
\usepackage[T1]{fontenc}    %
\usepackage{hyperref}       %
\usepackage{url}            %
\usepackage{booktabs}       %
\usepackage{amsfonts}       %
\usepackage{nicefrac}       %
\usepackage{microtype}      %
\usepackage{stackengine}
\usepackage{epigraph}
\usepackage{xcolor,soul}
\usepackage{algorithm, mathtools}
\usepackage{algorithmic}

\usepackage{adjustbox}

\usepackage[toc,page]{appendix}

\setstackgap{S}{7pt}
\setstackgap{L}{7pt}
\title{Model-Agnostic Learning to Meta-Learn}

  \author{\Name{Arnout Devos\nametag{\textsuperscript{\normalfont 1}\thanks{Denotes equal contribution.}}} \Email{arnout.devos@epfl.ch}\and
   \Name{Yatin Dandi}\nametag{\textsuperscript{\normalfont 1,}\textsuperscript{\normalfont 2}\footnotemark[1]} \Email{yatind@iitk.ac.in}\\
   \addr \textsuperscript{1}School of Computer and Communication Sciences, EPFL, Switzerland\\ \textsuperscript{2}Department of Computer Science and Engineering, IIT Kanpur, India}

\definecolor{applegreen}{rgb}{0.55, 0.71, 0.0}
\definecolor{ao(english)}{rgb}{0.0, 0.5, 0.0}
\definecolor{blue-violet}{rgb}{0.54, 0.17, 0.89}

\usepackage{amsmath}

\newcommand{\task}{\mathcal{T}}
\newcommand{\loss}{\mathcal{L}}

\newcommand{\learner}{f}

\newcommand{\lossi}{\loss_{\task_i}}

\newcommand{\lossc}{\loss_{\task_c}}

\begin{document}

\maketitle

\begin{abstract}
In this paper, we propose a learning algorithm that enables a model to quickly exploit commonalities among related tasks from an unseen task distribution, before quickly adapting to specific tasks from that same distribution.
We investigate how learning with different task distributions can first improve adaptability by meta-finetuning on related tasks before improving goal task generalization with finetuning.
Synthetic regression experiments validate the intuition that learning to meta-learn improves adaptability and consecutively generalization. Experiments on more complex image classification, continual regression, and reinforcement learning tasks demonstrate that learning to meta-learn generally improves task-specific adaptation.
The methodology, setup, and hypotheses in this proposal were positively evaluated by peer review before conclusive experiments were carried out.
\end{abstract}


\section{Introduction}

Recent years have seen encouraging developments in meta-learning based approaches for deep neural networks and their successful application to various domains \citep{MAML,imaml,nichol2018firstorder}. These approaches typically assume a distribution over tasks and aim to exploit the shared properties across tasks to learn a model that can adapt to unknown tasks from this distribution using only a few training data points.
Unfortunately, their adaptive capabilities do not generalize well to unseen tasks from related but different task distributions \citep{CloserLook}.

A number of recent works have proposed addressing the presence of different sets of related tasks by explicitly factoring in the heterogeneous nature of the task distribution in the design of the architecture and update rule \citep{CNAPs,ARML,HSML,mmaml}. However, these approaches still assume a fixed task distribution, such as  tasks sampled from a fixed set of families of functions or a multi-modal distribution arising out of a fixed set of task datasets. We argue that generalizing across unknown datasets and task distributions is a fundamentally more difficult problem than fixed distribution meta-learning. With a new task distribution or dataset, it is unrealistic to expect the model to quickly adapt to any arbitrary task from such a distribution. Instead, we expect the model to quickly learn to adapt to any task from the new task distribution after being exposed to only a few tasks of it. This generalizes the notion of few-shot learning to \textit{few-task} (few-shot) learning. Changes in task distributions might also arise due to natural or artificial transformations of the data. With different task distributions arising from a "distribution over task distributions", it is not only desirable to "quickly adapt" to unseen tasks but also "quickly learn to adapt" to unseen task distributions.

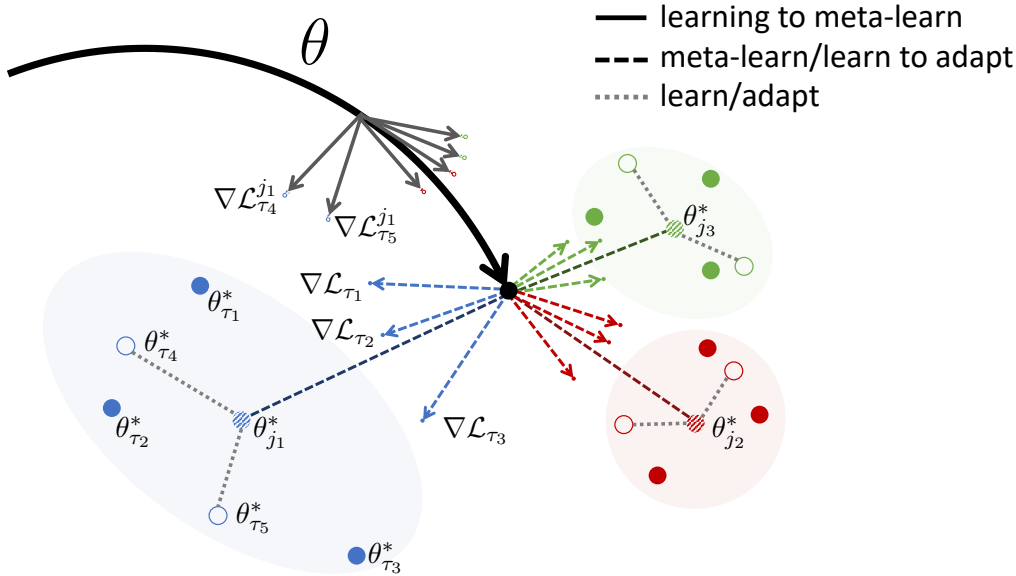
\begin{figure}[htbp]
\setlength{\unitlength}{0.5\columnwidth}
\begin{picture}(1.99,0.93) \linethickness{0.5pt}
\put(0,0){\includegraphics[width=\textwidth]{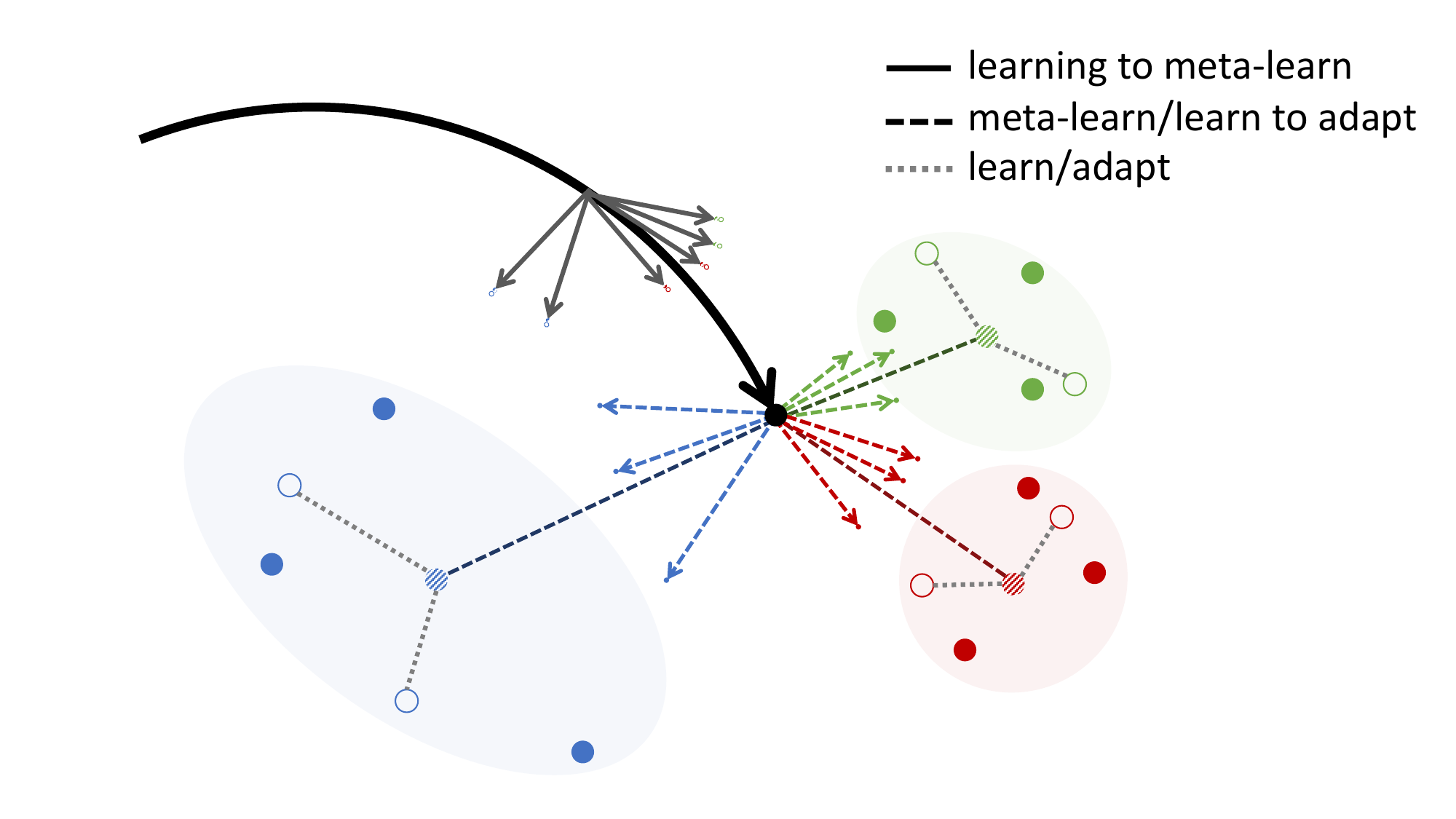}}
\put(0.7,0.95){\Huge$\theta$}
\put(0.55,0.7){$\nabla \mathcal{L}^{j_{1}}_{\tau_4}$}
\put(0.76,0.65){$\nabla \mathcal{L}^{j_{1}}_{\tau_5}$}

\put(0.70,0.55){$\nabla \mathcal{L}_{\tau_1}$}
\put(0.72,0.47){$\nabla \mathcal{L}_{\tau_2}$}
\put(0.95,0.30){$\nabla \mathcal{L}_{\tau_3}$}
\put(0.62,0.3){$\theta^{*}_{j_{1}}$}
\put(0.54,0.52){$\theta^{*}_{\tau_1}$}
\put(0.38,0.3){$\theta^{*}_{\tau_2}$}
\put(0.82,0.08){$\theta^{*}_{\tau_3}$}
\put(0.43,0.45){$\theta^{*}_{\tau_4}$}
\put(0.59,0.15){$\theta^{*}_{\tau_5}$}
\put(1.42,0.3){$\theta^{*}_{j_{2}}$}
\put(1.37,0.67){$\theta^{*}_{j_{3}}$}
\end{picture}
\caption{
Illustration of our model-agnostic learning to meta-learn algorithm (MALTML), which optimizes for a representation $\theta$ that can quickly adapt to new task distributions $j$ and consecutively to their tasks $\tau$. Illustrated with single gradient steps for the meta-learning and learn/adapt phases.
\label{fig:maltmldiagram}
}
\end{figure}

We propose a general framework for adapting to unseen task distributions by "learning to meta-learn" on different task distributions during training. Thus, the heterogeneity of tasks in our approach is not fixed but flexibly modeled through hierarchical sampling from a distribution over task distributions. 
Similar to MAML \citep{MAML}, we propose a general framework to learn a suitable initialization for a single set of parameters. Unlike MAML, which only trains a model to quickly adapt the parameters on a new task using few task-specific gradient steps, our model is also trained to quickly adapt its initialization to a new task distribution using few meta gradient steps on this unseen task distribution (see Figure \ref{fig:maltmldiagram}). We hypothesize that our approach would allow models to transfer learn capabilities across datasets in supervised and unsupervised learning, and new environments in reinforcement learning as well as quickly adapt to unseen augmentations and distortions at test time.

\section{Related Work}

Model-Agnostic Meta-Learning (MAML) by \citet{MAML} is a seminal work in few-shot meta-learning which seeks a common model initialization that allows the model to perform well on any goal task from the training task distribution with few gradient steps (and samples).
Multimodal MAML (MMAML) by \citet{vuorio2018multimodal, mmaml} extends MAML with the capability to identify tasks sampled from a multimodal task distribution and adapt quickly through gradient updates. \citet{HSML} proposed the hierarchically structured meta-learning (HSML) algorithm that explicitly tailors the transferable knowledge to different clusters of tasks. Automated Relational Meta-Learning (ARML) by \citet{ARML} extracts the cross-task relations and constructs a meta-knowledge graph. When a new task arrives, it can quickly find the most relevant structure and tailor the learned structure knowledge to the meta-learner. Still, MMAML, HSML, and ARML only learn to learn from a \textit{fixed} task distribution. Unlike our approach, they are not expected to generalize to new task distributions.

Research on improving meta-learning algorithms is vast, and we will highlight work most related to our approach. Following MAML, \citet{MetaSGD} and \citet{antoniou2019train} learn the inner learning rate in the outer loop to improve performance, while reducing the hyperparameter tuning requirement.
\citet{MAML} proposed a first-order approximation of MAML (fo-MAML) to scale to larger models, which was subsequently improved upon by \citet{nichol2018firstorder} with a first-order method called Reptile. 
Reptile can naturally be extended to our proposed approach, making it scalable and efficient. \citet{chen2020new} found that with increasingly deep architectures, common pre-training and transfer learning can outperform meta-learning from scratch in the visual classification domain. Based on this, they proposed to combine regular pre-training with subsequent meta-learning, which empirically gives a further performance improvement. \citet{raghu2020rapid} found that feature reuse is a dominant factor in MAML, and proposed a variant called Almost no Inner Loop (ANIL) which learns to only fine-tune the last layer linear classifier. On the contrary, \citet{oh2020does} came to the opposite conclusion that fast learning is crucial, and proposed a Body Only update in Inner Loop (BOIL) algorithm. These works motivate several of our research questions for the experiments.

\section{Methodology}
We aim to train models that can quickly change their adaptability before rapid adaptation. We formalize this setting as few-task few-shot learning. In this section, we will clarify the problem setup and we will formalize this learning to meta-learn problem setting for supervised learning, but it can easily be generalized to unsupervised and reinforcement learning (RL).
\subsection{Learning to Meta-Learn Problem Setup}

In our learning to meta-learn scenario, we consider a distribution $p(j)$ over task distributions such as families of related functions or datasets with similarities. Our goal is to allow the model to adapt to unseen task distributions as well as specific tasks within such distributions. In the $L$-task $K$-shot setting, the model is trained to meta-learn a new task distribution $j_d$ from only $L$ tasks with only $K$ examples each for task-learning and $Q$ examples for meta-learning, before learning a single goal task $\task_i$ drawn from $p_{j_d}(\task)$ from only $K$ samples. The model $f$ is then improved by considering how the test error on new (validation) data from $\task_i$ changes with respect to the original parameters. This test error on the final goal tasks serves as the training error of the learning to meta-learn process. At the end of training, new families are sampled from $p(j_d)$ and the model's learning to meta-learn performance is measured by the model's performance after meta-finetuning on $L$ tasks with $K+Q$ examples and finetuning on one or multiple goal tasks from the same family with $K$ examples.

\subsection{A Model-Agnostic Learning to Meta-Learn Algorithm}

We propose a method that can learn the parameters of any model via learning to meta-learn in such a way as to prepare that model to first quickly change its adaptation capability (initialization) and consecutively adapt quickly to a goal task. The intuition behind this approach is that some internal representations are more transferable to meta-learn with. For example, a neural network could learn features that are broadly applicable to all tasks in the distribution over task distributions $p(j)$ and can then specialize to an individual task distribution $p_{j_d}(\task)$, rather than to a single task distribution or task. We assume no specific form of the model, other than that it is parametrized by some parameter set $\mathbf{\theta}$, and that the loss functions are sufficiently smooth in $\theta$ such that we can employ gradient-based learning techniques.

Formally, we consider a model represented by a parametrized function $\learner_\theta$ with parameters $\theta$. When adjusting the adaptability to a new task \textit{family} $j_d$, the model's parameters $\theta$ become $\theta_{j_d}'$, and consecutively when adapting (finetuning) to a new task $\task_c \sim p_{j_d}(\task)$ the model's parameters $\theta_{j_d}'$ become $\theta_c'$.
In our approach, the updated parameter vector $\theta_{j_d}'$ is obtained by few (inner) \textit{meta}-learning steps on few tasks from dataset $j_d$, also called \textit{meta-finetuning}.
Each meta-finetuning step is taken across few task-finetuning steps.
A single ($r=1$) \textit{task} gradient step with step size $\alpha$ is:

\begin{equation}\label{eq:sgd}
    \theta_{\task_i}'= U^{r=1, \alpha}_{\task_i} (\theta) = \theta-\alpha \nabla_\theta  \lossi (\learner_{\theta}).
\end{equation}

Then, the family-specific meta-finetuning update, with one ($r=1$) task-level parameter update as in Equation \eqref{eq:sgd} and one ($m=1$) meta-level update across tasks from $j_d$ with step size $\beta$, is:
\begin{equation}
    \theta_{j_d}'= V^{m=1, \beta}_{j_d} (\theta) = \theta-\beta \nabla_\theta  \left (\sum_{\task_i \sim p_{j_d}(\task)}  \lossi ( \learner_{U^{r=1, \alpha}_{\task_i} (\theta)})\right ).
\end{equation}

Consecutively, the updated task-specific parameter vector $\theta_c'$ is obtained by taking few gradient descent steps, with learning rate $\gamma$, on tasks $\task_c \sim p_{j_d}(\task)$, starting from $\theta_{j_d}'$. For example, using Equation \eqref{eq:sgd}, with 1 gradient step: $\theta_c' = U^{r=1,\gamma}_{\task_c} (\theta_{j_d}')$. Note that using the same $r$ for meta-finetuning and goal task finetuning is a natural choice, but can be deviated from. The step-sizes $\alpha, \beta$, and $\gamma$ may be fixed as hyperparameters or learned on the outer learning loop (see below).

Finally, the global model parameters $\theta$ are optimized to perform well after this two-step (meta-learn, then learn) process. Specifically, the model parameters are trained by optimizing the performance of every $f_{\theta_c'}$ on its task $\task_c$. This is done across task distributions sampled from $p(j)$ and tasks sampled from them ($\task_c \sim p_{j_d}(\task))$. Concretely, the \textit{learning-to-meta-learn} objective is:

\begin{align*}
\min_{\theta} \sum_{j_d \sim p(j)}\sum_{\task_c \sim p_{j_d}(\task)}  \lossc ( \learner_{\theta_c'}) = \min_{\theta} \sum_{j_d \sim p(j)}\sum_{\task_c \sim p_{j_d}(\task)}  \lossc \left( U^{r, \gamma}_{\task_c} (V^{m,\beta}_{j_d} (\theta))\right)\\
\end{align*}

Note that the learning to meta-learn optimization is performed over the model parameters $\theta$, whereas the learning to meta-learn objective itself is computed using the updated model parameters $\theta_c'$.

The outer optimization across task distributions is performed via stochastic gradient descent (SGD), such that the model parameters $\theta$ are updated as follows:

\begin{equation}
    \theta \leftarrow \theta - \eta \nabla_\theta \sum_{j_d \sim p(j)}\sum_{\task_c \sim p_{j_d}(\task)}  \lossc ( \learner_{\theta_c'})
\end{equation}

where $\eta$ is the outer step size. The full algorithm, in the general case, is outlined in Algorithm \ref{alg:maltml}.

The MALTML outer gradient update yields a third-order gradient with respect to $\theta$. To make MALTML computationally usable for high-dimensional models, we propose a first-order approximation. Concretely, following \citet{nichol2018firstorder}, for every family $j_d$ we employ multiple \textit{first-order} meta-learning updates $\theta_{j_d}' = \widetilde{V}^{m>1}_{j_d}(\theta)$, before updating the model parameters with $\theta \leftarrow \theta + \eta \sum_{j_d} (\theta_{j_d}' - \theta)$. Note that in this case, due to the nature of our two-level first-order approximation, goal task finetuning is not required anymore.

\begin{algorithm}[t]
\caption{Model-Agnostic Learning to Meta-Learn}
\label{alg:maltml}
\begin{algorithmic}[1]
{\footnotesize
\REQUIRE $p(j)$: distribution over task distributions parameter $j$
\REQUIRE $\alpha$, $\beta$, $\gamma$, $\eta$: step size hyperparameters
\STATE randomly initialize $\theta$
\WHILE{not done}
\STATE Sample batch of task distributions $j_d \sim p(j)$
\FORALL{$j_d$}
  \STATE Sample $L$ tasks $\task_i \sim p_{j_d}(\task)$
  \FORALL{$\task_i$} 
      \STATE Evaluate $\nabla_\theta \lossi(\learner_\theta)$ with respect to $K$ examples
      \STATE Compute adapted parameters with gradient descent: $\theta_i'=\theta-\alpha \nabla_\theta  \lossi(  \learner_\theta )$
 \ENDFOR
 \STATE Update $\theta_{j_d}' \leftarrow \theta - \beta \nabla_\theta \sum_{\task_i}  \lossi ( \learner_{\theta_i'})$ using $Q$ examples per task
 \STATE Sample batch of tasks $\task_c \sim p_{j_d}(\task)$
  \FORALL{$\task_c$}
      \STATE Evaluate $\nabla_{\theta_{j_d}'} \lossc(\learner_{\theta_{j_d}'})$ with respect to $K$ examples
      \STATE Compute adapted parameters with gradient descent: $\theta_c'=\theta_{j_d}'-\gamma \nabla_{\theta_{j_d}'}  \lossc(  \learner_{\theta_{j_d}'} )$
 \ENDFOR
 \ENDFOR
 \STATE Update $\theta \leftarrow \theta - \eta \nabla_\theta \sum_{j_d \sim p(j)}\sum_{\task_c \sim p_{j_d}(\task)}  \lossc ( \learner_{\theta_c'})$
\ENDWHILE
}
\end{algorithmic}
\end{algorithm}

\section{Experimental Protocol}
The goal of our experimentals is to get conclusive results in different learning domains on whether MALTML can enable a quick and significant change of adaptability to goal tasks from new task distributions. Moreover, we wish to examine whether fast learning (to meta-learn) \citep{oh2020does} or feature reuse \citep{raghu2020rapid} is the dominant factor in the performance.

All the learning to meta-learn problems we consider require some form of change in adaptability and subsequent adaptation to new tasks at test time. When possible, we will compare our results to an oracle that receives the identities of the family and goal task as an additional input or a MAML oracle that is able to meta-finetune on a large number of tasks from the new task distribution, as upper bounds on the performance of the models. Regarding model architecture and optimization, we will follow \citet{MAML}. We will use insights from \citet{antoniou2019train} to stabilize training where applicable and follow its hierarchical hyperparameter search methodology. We will carry out the experiments in PyTorch \citep{paszke2019pytorch}, using the torchmeta package \citep{deleu2019torchmeta}. The code will be available online.

\subsection{Illustrative preliminary experiment: regression}\label{sec:exp:regression}
We start with a toy regression problem which illustrates the experimental protocol of few-task few-shot learning and the basic principles of MALTML.

Each task distribution (family) consists of sinusoid regression tasks with a specific phase. Thus, $p(j)$ is continuous, where the phase $j$ varies uniformly within $[0, \pi]$. Each task involves regressing from the input to the output of a sine wave, where the amplitude is varied between tasks. Thus, $p_{j_d}(\task)$ is continuous, where the amplitude varies within $[0.1, 5.0]$ and the input and output both have a dimensionality of 1. During training and testing, datapoints $x$ are sampled uniformily from $[-5,5]$. The loss is the mean squared error between the prediction $f(x)$ and the true value. The model is a neural network with 2 hidden layers of size 40 with ReLU nonlinearities. When training with MALTML, we use single gradient updates with fixed step sizes of $\alpha=0.001$, $\beta=0.01$, $\gamma=0.001$, and use Adam \citep{kingma2017adam} with an initial learning rate of $\eta=0.001$. The baselines are also trained with Adam, and an inner learning rate of $\alpha=0.001$ for MAML. For the 5-task 5-shot regression experiment, we train for 70,000 outer steps with a family batch size of 10, $Q=5$, and 2 validation tasks per family.

For this preliminary experiment we only contrast with a MAML baseline, which disregards the family structure, and an oracle receiving the true amplitude and phase of the goal task as additional input. In general, we intend to compare to the other oracles described before and another baseline: pretraining on all tasks, which in this case involves training a model to regress random sinusoid functions.

\begin{figure*}[htbp]
   \subfigure{\label{fig:maltml_toy}%
      \includegraphics[width=0.32\textwidth]{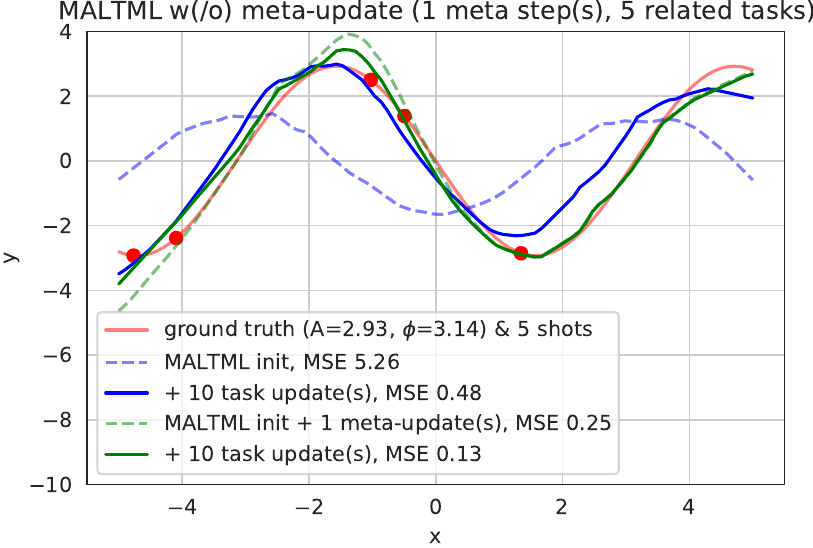}}
   \subfigure{\label{fig:maltml_toy_2}%
      \includegraphics[width=0.32\textwidth]{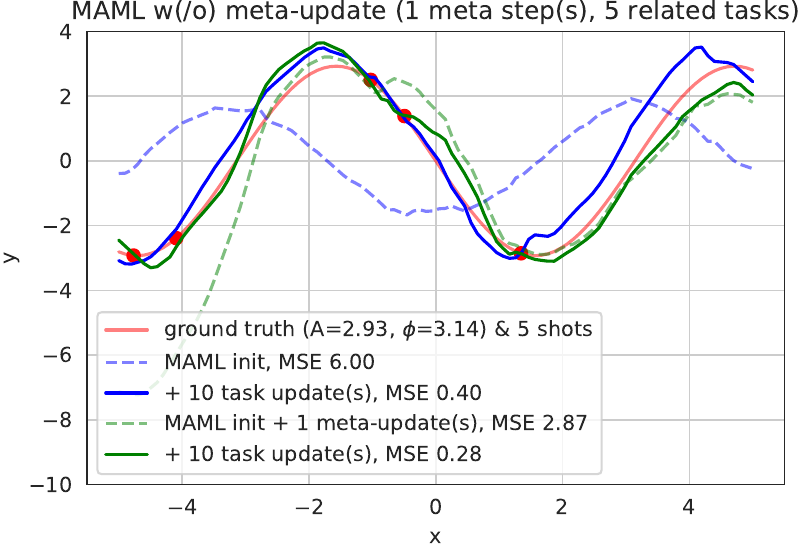}}
   \subfigure{\label{fig:maltmlvsmaml}%
      \includegraphics[width=0.32\textwidth]{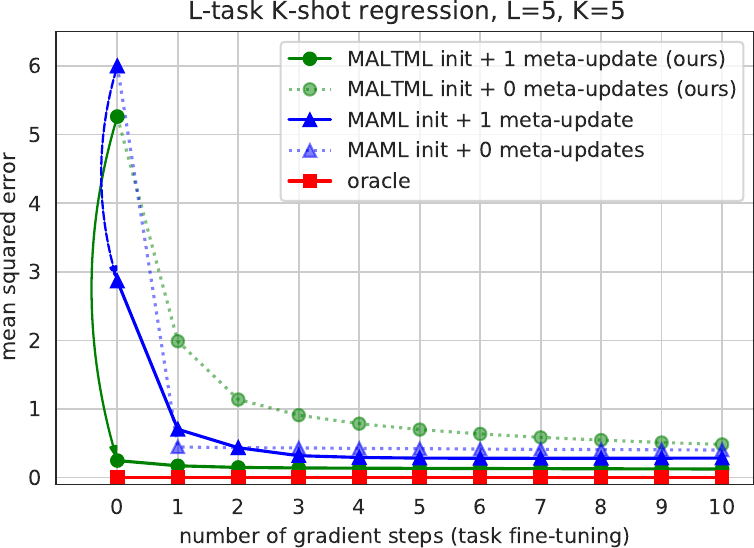}}
\vspace{-.4cm}
\caption{\label{workflow} Effect of the meta-update and task fine-tuning on (a) MALTML (b) MAML (c) Test error vs task fine-tuning steps (the meta-update improvement is indicated by the arrows at $0$ steps).}
\end{figure*}

The toy results in Figures \ref{fig:maltml_toy} and \ref{fig:maltmlvsmaml} show that the learned MALTML model is able to quickly change its adaptability to the new family's phase on 5 related 5-shot tasks. Due to this, it reaches a better fit than MAML, which benefits less from the meta-adaptation of its initialization (Figures \ref{fig:maltml_toy} and  \ref{fig:maltmlvsmaml}).

\subsection{Specifics of main experiments}
Besides the toy experiment in section \ref{sec:exp:regression}, we propose to test the effectiveness of MALTML for:
\paragraph{Classification.} We propose to apply our method to modified versions of the Omniglot \citep{lake2015human} and ImageNet \citep{ILSVRC15} datasets. For ImageNet, we will propose a few-task few-shot dataset, making use of its hierarchical structure to generate a sufficient amount of training families. Given the finite number of alphabets in Omniglot, which will serve as families, we will use data augmentations similar to the ones used in \citet{khodadadeh2019unsupervised} to generate a large number of training families. To arrive at a realistic setting where the (imposed) hierarchical structure from the supervised case is lacking, we will use a hierarchically augmented version of unsupervised meta-learning \citep{khodadadeh2019unsupervised}. Specifically, a subset of data augmentation parameters will be sampled per family, before applying them randomly (with the remaining subset) on samples to generate tasks from each family. Note that in this case, on a (family) meta-level the method needs to be sensitive to augmentations, whereas on a task-level it should aim to be invariant to them.
\paragraph{2D Navigation.} We propose to evaluate MALTML on a set of families of RL tasks where a point agent must move to different goal positions in 2D, while being given related tasks from the same family. Every family constitutes of a random crop of the unit square, and every task is randomly chosen from within that rectangle. The crops are bounded by $25\%$ to $75\%$ of the original unit length.
\paragraph{Continual Regression}We propose to evaluate a continual learning extension of MALTML on incremental sine wave learning as described in \citet{javed2019metalearning}. Different families of continual learning prediction problems will correspond to different frequencies of the sine waves. The inner meta-learning objective for this setting will be replaced by the Online-aware Meta-Learning objective from \citet{javed2019metalearning}.

\paragraph{Continual Reinforcement Learning} Besides the sine waves continual regression problem, we will aim to evaluate a more challenging and real-world setting of continuous control inspired by \citet{kaplanis2020continual}.

\section{Future Work}

Based on the results of our main experiments, future work could involve a multi-task setting corresponding to meta-learning on different categories of tasks such as classification, segmentation, and depth estimation on a single dataset or a set of related datasets. This could also involve augmenting our objective for enforcing cross-task (distribution) consistency \citep{zamir2020robust}.

\section{Results}
\subsection{Experimental Setup and Hyperparameters}

For the tasks included in \citet{MAML}, i.e., Omniglot classification, 2D Navigation and Half-Cheetah goal velocity tasks, we use the same topmost optimizer and learning rate, fine-tuning step size, number of gradient steps, and number of tasks for the meta-step for the MAML baselines and the inner task specific updates for MATML.  For the other baselines and tasks, we borrow the architectures and hyperparameters from works that compare with MAML's original setup, i.e., \citet{raghu2020rapid} for ANIL and \citet{oh2020does} for BOIL and tiered-ImageNet. The used architectures and hyperparameters are further described in Appendix \ref{appendix:omniglot_supp} and \ref{appendix:tiered_supp}. For the inner meta-step size, we performed a grid search over the set of values $\beta \in \{2,1,0.5,0.1,0.05,0.01,0.001,0.0001\}$. Due to computational constraints on our Tesla V100-SXM2 32GB GPU, for tiered-ImageNet and continual regression, we sample only one family per outer update. Other details specific to the tasks are discussed in the corresponding sections and the Appendix.
\subsection{Classification}

\subsubsection{Omniglot}\label{sec:omniglot_results}
Although the original Omniglot dataset \citep{lake2015human} provided in characters from separated alphabets for training and testing, it did not provide in a pre-defined validation set as is common in few-shot learning \citep{vinyals2017matching}. In \citet{vinyals2017matching} different sets of characters for training, validation and testing are sampled, disregarding the alphabet structure. Needing to separate alphabets across these sets for learning to meta-learn, we created new sets. The details of the adapted dataset are described in Appendix \ref{appendix:omniglot_supp}. Since we use a different dataset organization than most other literature, we reproduce results for the baselines as well. Specifically, we reproduce results for MAML \citep{MAML}, ANIL \citep{raghu2020rapid} and BOIL \citep{oh2020does} for an honest comparison of rapid learning and feature reuse. Our equivalent learning to meta-learn methods are MALTML, MALTML-ANIL, and MALTML-BOIL, respectively. Appendix Table \ref{tab:omniglot_hyperparameters} lists the hyperparameters used for Omniglot. Table \ref{tab:omniglot_results} shows the results on Omniglot.
\begin{table}[hbtp]
\floatconts
  {tab:Omniglot}
  {\caption{Accuracies of 5-way (5-task) Few-shot learning on 16 held-out Omniglot alphabets, before and after adaptation to the test families. Includes $95 \%$ confidence intervals across 600 test tasks from the 16 test alphabets. %
  }\label{tab:omniglot_results}}
  {\hspace*{-.75cm}
  \begin{tabular}{lllll}
  \toprule
  \bfseries Model & \bfseries 1-shot  & \bfseries 5-task 1-shot & \bfseries 5-shot  & \bfseries 5-task 5-shot  \\
  \midrule
   MAML \citep{MAML} & 97.86 $\pm$ 0.29\% & 98.04 $\pm$ 0.30\%  & 99.32 $\pm$ 0.17\% & 99.26 $\pm$ 0.19\% \\ %
   ANIL \citep{raghu2020rapid} & 96.50 $\pm$ 0.41\% & 96.86 $\pm$ 0.38\%  & 98.62 $\pm$ 0.27\% & 98.60 $\pm$ 0.26\% \\ %
   BOIL \citep{oh2020does} & 97.39 $\pm$ 0.35\% & 97.57 $\pm$ 0.32\%  & 99.10 $\pm$ 0.22\% & 99.19 $\pm$ 0.18\% \\ %
   MALTML (ours) & 95.82 $\pm$ 0.41\% & 96.49 $\pm$ 0.35\%  & 98.65 $\pm$ 0.21\% & 98.77 $\pm$ 0.19\% \\ %
   MALTML-ANIL (full, ours) & 91.30 $\pm$ 0.69\% & 92.73 $\pm$ 0.65\%  & 94.92 $\pm$ 0.56\% & 95.12 $\pm$ 0.58\% \\ %
   MALTML-BOIL (full, ours) & 97.04 $\pm$ 0.35\% & 97.28 $\pm$ 0.32\%  & 98.85 $\pm$ 0.23\% & 99.20 $\pm$ 0.17\% \\ %
  \bottomrule
  \end{tabular}}
\end{table}

\subsubsection{Tiered-ImageNet}

In a more challenging setting, we evaluate MATLML and the baselines on the Tiered-Imagenet dataset with the families corresponding to different categories of classes, as defined in \citet{ren2018metalearning}. Following \citet{ren2018metalearning}, the 34 categories are divided into 20 training, 6 validation, and 8 test categories. We visualize a selection of categories (families) and their tasks in Appendix \ref{appendix:tiered_supp}. Table \ref{tab:tiered_results} shows the results on tiered-ImageNet.

\begin{table}[hbtp]
\floatconts
  {tab:Tiered Imagenet}
  {\caption{5-way Few-shot and 5-way Few-task Few-shot classification on held-out tiered-ImageNet families, before and after adaptation to the test families. The $\pm$ shows $95 \%$ confidence intervals over families. Here MALTML-Reptile refers to the model utilizing the first-order approximation for the inner meta-steps. MALTML-ANIL (partial) and MALTML-ANIL (full) refer to the models updating only the head in task specific adaptation and task as well as family specific adaptation, respectively. Similarly, MALTML-BOIL (partial) and MALTML-BOIL (full) refer to updating only the body in the corresponding settings. The size of the confidence intervals is relatively large due to the limited number of 8 test families. }\label{tab:tiered_results}}
  {\hspace*{-.75cm}
  \begin{tabular}{lllll}
  \toprule
  \bfseries Model & \bfseries 1-shot  & \bfseries 4-task 1-shot & \bfseries 5-shot  & \bfseries 2-task 5-shot  \\
  \midrule
   MAML \citep{MAML} & $46.3 \pm 2.4\%$ & $46.5 \pm 2.8\%$ & $62.1 \pm 2.6\%$ & $62.6 \pm 3.2\%$\\
   ANIL \citep{raghu2020rapid} & 47.2 $\pm$ 1.3\% & 46.4 $\pm$ 1.9\%  &  62.8 $\pm$ 2.2\% & 63.1 $\pm$ 2.4\% \\ 
   BOIL \citep{oh2020does} & 48.4 $\pm$ 1.2\% & 48.9 $\pm$ 2.4\%  & 65.7 $\pm$ 2.4\% & 66.4 $\pm$ 3.1\% \\ 
   MALTML (ours) &$45.4 \pm 1.6\%$ & $47.2 \pm 3.2\%$  & $60.8 \pm 3.4\%$ &$63.9 \pm 4.1\%$\\
   MALTML-Reptile (ours) & $32.9 \pm 2.4\%$ & $33.6 \pm 2.9\%$ & $48.3 \pm 3.2\%$& $49.1 \pm 3.5\% $\\
   MALTML-ANIL (partial, ours)& $34.4 \pm 1.4\%$ & $38.3 \pm 2.4\%$ & $51.5 \pm 1.7\%$ & $54.7 \pm 2.2\%$\\
   MALTML-ANIL (full, ours)  & $34.1 \pm 1.9\%$ & $39.5 \pm 2.6\%$& $ 52.3 \pm 2.0\% $ & $52.8 \pm 2.9\%$ \\
   MALTML-BOIL (partial, ours) & $37.6\pm 1.8\%$ & $39.2 \pm 2.5\%$ & $55.4 \pm 2.3\%$ & $57.3 \pm 3.7\%$\\
   MALTML-BOIL (full, ours) & $37.3 \pm 2.0\%$ & $38.8 \pm 3.1\%$ & $56.1 \pm 2.4\%$ & $58.8 \pm  3.8\%$\\
  \bottomrule
  \end{tabular}}
\end{table}

\subsection{2D Navigation}
Inspired by the 2D navigation problem in \citet{MAML}, we evaluate an equivalent family-tuning before task-tuning setting. Specifically, before goal task fine-tuning, we give the agent few related tasks to meta-tune on. All tasks are sampled from a box region of $x\in(-.5,.5)$ and $y\in(-.5,.5)$. A family with lower bound $a$ and upper bound $b$ constitutes of tasks with goals within the area $x\in(a, b)$ and $y\in(a, b)$. $a$ and $b$ are randomly sampled from $[-0.5,0.5]$. We use the same hyperparameters as in \citet{MAML}, a meta-learning rate of $0.01$, an outer training batch size of 1, and 20 support tasks and 20 validation tasks.
To allow large changes in the policy due to meta updates, we use policy gradient for both inner task specfic updates and inner meta-steps, while the topmost updates are obtained using TRPO \citep{pmlr-v37-schulman15}. 
Appendix \ref{appendix:2d_results} provides illustrations. Table \ref{tab:2DRL_results} shows the results on the 2D navigation problem.

\begin{table}[hbtp]
\floatconts
  {tab:2d}
  {\caption{Average return for (meta-)adapting the 2D Navigation RL policy. Higher return is better, with 0 being the maximal reward.}\label{tab:2DRL_results}}
  {\hspace*{-1.5cm}
  \begin{tabular}{lll|ll}
  \toprule
  \bfseries Model & \bfseries init(ialization)  & \bfseries 1 task update & \bfseries 1 meta-update init & \bfseries 1 task update  \\
  \midrule
   MAML \citep{MAML}  & -12.6 & -11.7 & -13.6 & -11.3\\
   MALTML (ours) & -38.4 & -27.7 & -13.3 & -10.2 \\
  \bottomrule
  \end{tabular}}
\end{table}

\subsection{Continual Regression}

\begin{figure*}[t]
    \subfigure{\label{fig:OML_regression}%
      \includegraphics[width=\textwidth]{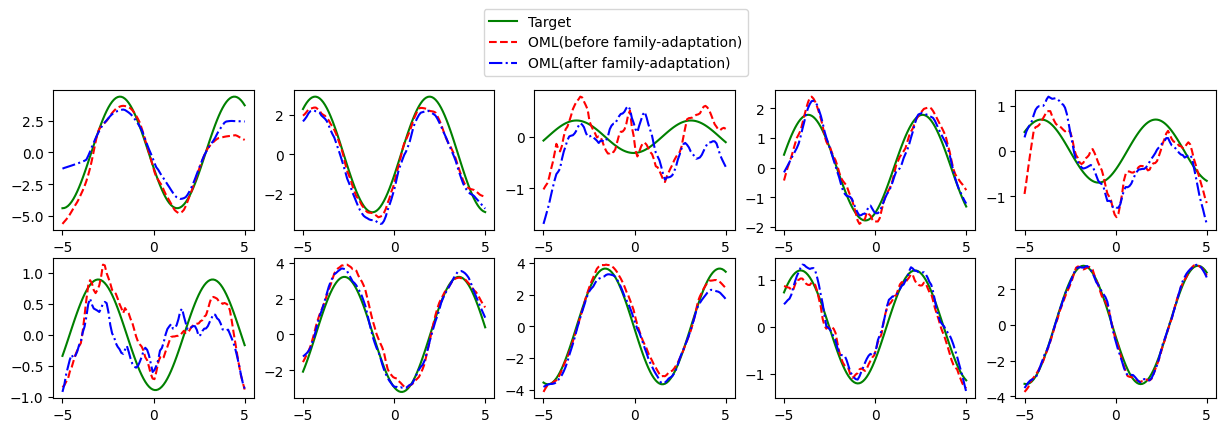}}
    \subfigure{\label{fig:OMALTML_regression}%
        \includegraphics[width=\textwidth]{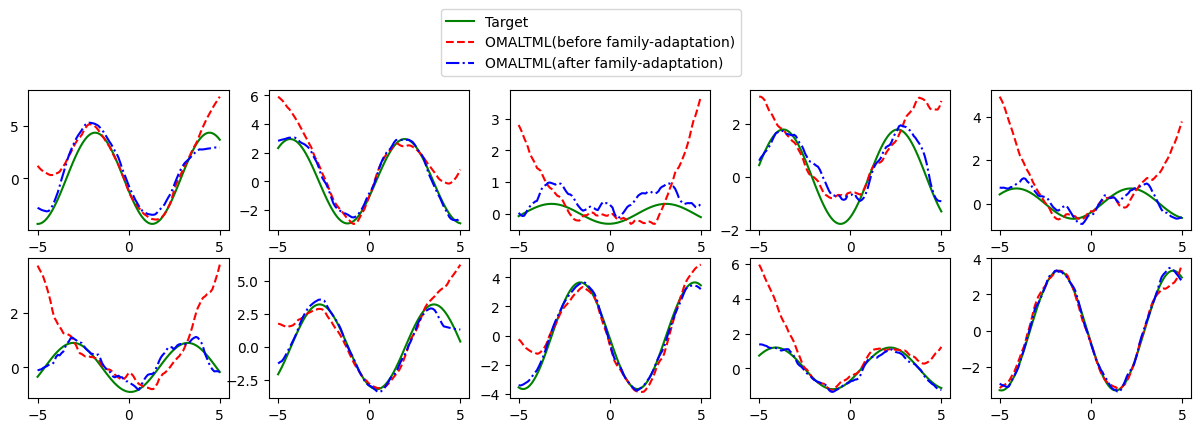}}
\caption{\label{fig:few_shot_cont_reg} Few-Task Few-shot adaptation in a continual regression task for a randomly sampled trajectory before and after adaptation to another trajectory from the same family: (a) MAML (b) MALTML. Note that MALTML's ability to adapt to the trajectory signficantly improves due to the meta-step while MAML's output function demonstrates negligible change.}
\end{figure*}

Following \cite{javed2019metalearning}, we extend the few-shot regression task to continual learning over trajectories of tasks consisting of sequences of sinusoid functions of randomly sampled frequencies, amplitudes, and phases. The families (task distributions) are constructed by using the same randomly sampled frequency for each task (trajectory) within a family. Further details about the architecture, hyperparameters, and the sampling procedure are provided in Appendix \ref{appendix:cont_reg}. We refer to the online extensions of MALTML and MAML as OMALTML and OML, respectively. To ensure fair comparison, we utilize the same architecture and inner loop updates as  \citet{javed2019metalearning}, who split the training network into representation learning and prediction learning modules, with only the prediction learning modules being updated in the inner loop. 
Figures \ref{fig:contreg_mse}, \ref{fig:OML_regression}, and \ref{fig:OMALTML_regression} demonstrate the effectiveness of OMALTML in adapting to unseen families of trajectories to improve the subsequent task specific adaptation on the given family.
\begin{figure}[t]
\floatconts
  {fig:contreg_mse}
  {\caption{Mean squared error for OMALTML and OML averaged over 50 sinusoid families (frequencies) with $95\%$ confidence interval drawn by 1,000 bootstraps. The x-axis denotes the number of functions in a given trajectory having being utilized to provide task(trajectory) specific updates to the model.}}
  {\includegraphics[width=0.5\linewidth]{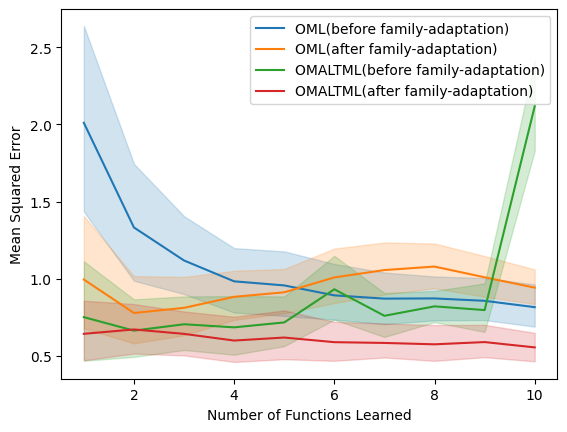}}
\end{figure}

\subsection{Continual Reinforcement Learning}

We evaluate MALTML's adaptation capabilities on continual Reinforcement Learning through the half-cheetah locomotion task with oscillating gravity recently introduced by \citet{kaplanis2020continual}. Each family is constructing by adding an oscillating function to a baseline gravity value of $-12$. The oscillating function corresponds to a sinusoid function with a phase, frequency and amplitude sampled uniformly from the ranges $[0,2.0],[0.1,5.0],[0,\pi]$ which describes the evolution of the environment's gravity with time.  While obtaining the gravity's value, the timesteps are divided by the horizon (set to 200), to scale the input range to $[0,1]$. Within each family, different tasks correspond to different goal velocities. Following \citet{kaplanis2020continual}, the gravity's value is appended to the input state at each timestep. Table \ref{tab:continual_results} shows the continual RL results.

\begin{table}[hbtp]
\floatconts
  {tab:Half-Cheetah}
  {\caption{Continual Reinforcement Learning: average return for Half-Cheetah with Oscillating Gravity RL policy. Higher return is better.}\label{tab:continual_results}}
  {\hspace*{-2cm}
  \begin{tabular}{lll|ll}
  \toprule
  \bfseries Model & \bfseries init(ialization)  & \bfseries 1 task update & \bfseries 1 meta-update init & \bfseries 1 task update  \\
  \midrule
   MAML \citep{MAML}  & -145.5 & -74.8 & -113.1 & -70.0\\
   MALTML (ours) 
   &-156.9 & -87.8  & -113.9 & -62.8  \\
  \bottomrule
  \end{tabular}}
\end{table}
\section{Findings}
\subsection{Rapid Meta-Learning}
 
 \subsubsection{MALTML}
As demonstrated through the results in Figure \ref{fig:contreg_mse}, and Tables \ref{tab:2d},\ref{tab:Half-Cheetah}, MALTML successfully learns to adapt to new task distributions using a meta step on a few tasks for continual regression. However, the improvements due to test time adaptation to new families is significantly limited for classification tasks (Tables \ref{tab:omniglot_results}, \ref{tab:Tiered Imagenet}). We hypothesize that this is primarily due to the limited number of training families in realistic image classification datasets unlike the continuous distribution over families available for continual regression.
Moreover, we found that the magnitude of improvement due to the meta-step is quite sensitive to hyperparameters such as the step size used for the meta-step during training. 
 
 \subsubsection{MAML}
 Through the results in Tables \ref{tab:omniglot_results},\ref{tab:tiered_results},\ref{tab:2d},\ref{tab:Half-Cheetah}, and Figure \ref{fig:contreg_mse}, we observe that unlike MALTML, MAML is unable to leverage the meta-steps on unseen tasks to improve adaptability on test task distributions. Thus training MALTML to quickly adapt through meta-steps is beneficial for adaptation to unseen task distributions.

\subsection{Rapid (Meta-)Learning is More Important than Feature Reuse}
The Omniglot results in Table \ref{tab:omniglot_results} and the tiered-ImageNet results in Table \ref{tab:tiered_results} show that \textit{rapid learning} is a dominant factor in achieving good classification accuracy. Specifically, for learning to meta-learn on Omniglot, there is a clear gap in performance of 2\% to 3\% between the setting where only the features are trained to rapidly meta-learn and learn (MALTML-BOIL) and the setting where only the classifier is trained to rapidly meta-learn and learn (MALTML-ANIL). Note however, that although \textit{rapid meta-learning} (using few related tasks) does seem to bring an overall performance improvement, it is not significant on Omniglot. For meta-learning, the performance improvement of BOIL over ANIL is not significant, but still consistently present, in agreement with the experiments in \citet{oh2020does}.

\subsection{Overfitting}
Through the tiered-ImageNet results in Table \ref{tab:Tiered Imagenet}, we observe that even though MALTML achieves significant improvement through the meta-step, the mean accuracy on test families can still be significantly lower than training task distribution top accuracies. We hypothesize that this occurs due to the model overfitting on the small number of training families (20 for tiered-ImageNet).
For a continuous distribution over families, such as in continual regression, MALTML obtains significant improvement over the baselines in adapting to unseen families. 

\subsection{Reptile}

As shown in Table \ref{tab:Tiered Imagenet}, using the first order Reptile \citep{nichol2018firstorder} approximation for the inner meta-steps leads to a significant drop in the performance. This suggests a need to design more effective first order approaches for the few-task few-shot learning task.
\section{Documented Modifications}\label{sec:documented_modifications}

\begin{enumerate}
    \item Instead of creating a custom hierarchical dataset from all the classes defined in Imagenet, we directly utilized the hierarchy introduced in tiered-ImageNet \citep{ren2018metalearning} based on 34 categories defined on 608 classes (779,165 images). This was done to ensure computational feasibility of the experiments and the absence of any baselines for meta-learning on the full Imagenet dataset.
    \item Contrary to the originally proposed oracle baselines, we primarily used  MAML as a baseline, since we found that the number of families and the number of tasks within each family were insufficient to obtain reliable family specific or adaptation based oracles. Thus, following the literature, e.g., \citep{raghu2020rapid, oh2020does}, we don't include the oracles, and focus on contrasting our approach with real-world baselines.
    \item We adjusted the 2D navigation task to have families defined by a single upper and lower bound. This eased the sampling of families and tasks.
    \item Self-Supervised Learning: upon deeper investigation based on the proposal, it has been determined that few-shot self-supervised learning using augmentations is not much closer to a real-life setting than the fully supervised case. Namely, its performance is very dependent on the augmentations used. These augmentations can only be tuned on a validation set with labels. To get to an adequate set of augmentations, one still needs a validation set consisting of (many) labeled validation families and tasks. This can be seen, e.g., in \cite{khodadadeh2019unsupervised}, where the augmentations are tuned on the Omniglot test set. Hence, we have not conducted this set of experiments.
\end{enumerate}

\section*{Acknowledgements}
We would like to express gratitude to the anonymous reviewers, as well as Matthias Grossglauser for insightful comments. Arnout Devos acknowledges funding from the European Union’s Horizon 2020 research and innovation program under the Marie Skłodowska-Curie grant agreement No. 754354.

\bibliography{main}
\newpage
\appendix

\section{Omniglot} \label{appendix:omniglot_supp}
Omniglot consists of 1623 handwritten characters from 50 alphabets and 20 examples per character. Identical to \citet{vinyals2017matching}, the grayscale images are resized to 28x28. However, to not have overlapping alphabets between training and testing sets, we resample the characters and alphabets according to Section \ref{sec:omniglot_alphabets}.
\subsection{Hyperparameters}
Table \ref{tab:omniglot_hyperparameters} provides the hyperparamters used for Omniglot training and testing.
\begin{table}[hbtp]
    \centering
    \caption{Omniglot hyperparameter summary.}
    \hspace*{-4em}\begin{tabular}{l r r r | r r r}
        \toprule
        &\multicolumn{3}{c}{Few-shot} & \multicolumn{3}{c}{Few-task Few-shot} \\
         Hyperparameter & MAML & ANIL & BOIL & MALTML & MALTML-ANIL & MALTML-BOIL \\
         \midrule
         Model architecture & Conv-4 & Conv-4 & Conv-4 & Conv-4 & Conv-4 & Conv-4\\
         Image input size & $28\times28$ & $28\times28$& $28\times28$& $28\times28$& $28\times28$& $28\times28$\\
         Outer optimizer & Adam & Adam & Adam & Adam & Adam & Adam\\
         Outer step size & 0.001 & 0.001 & 0.001 & 0.001 & 0.001 & 0.001 \\
         Query examples/task & 15 & 15 & 15 & 15 & 15 & 15\\
         Support Tasks/family  & n.a. & n.a. & n.a.  & 5 & 5 & 5\\
         Query Tasks/family  & n.a. & n.a. & n.a.  & 15 & 15 & 15\\
         Training batch size  & 16 tasks & 16 tasks & 16 tasks & 4 families & 4 families & 4 families\\
         \midrule
         Meta-tuning optimizer  & n.a. & n.a. & n.a. & SGD & SGD & SGD\\
         Meta-tuning step size  & n.a. & n.a. & n.a. & 0.4 & 0.1 & 0.4\\
         Meta-tuning steps  & n.a. & n.a. & n.a. & 1 & 1 & 1\\
         Meta-tune last layer  & n.a. & n.a. & n.a.  & \checkmark & \checkmark & \\
         Meta-tune backbone & n.a. & n.a. & n.a. & \checkmark & & \checkmark\\
         \midrule
         Fine-tuning optimizer  & SGD & SGD & SGD & SGD & SGD & SGD\\
         Fine-tuning step size  & 0.1 & 0.1 & 0.1 & 0.4 & 0.1 & 0.4\\
         Fine-tuning steps  & 1 & 5 & 1 & 1 & 5 & 1\\
         Fine-tune last layer  & \checkmark & \checkmark &  & \checkmark  & \checkmark & \\
         Fine-tune backbone & \checkmark  &  & \checkmark & \checkmark  &  & \checkmark \\
        \bottomrule
    \end{tabular}
    \label{tab:omniglot_hyperparameters}
\end{table}

\subsection{training (30), validation (4), test (16) alphabets}\label{sec:omniglot_alphabets}

\paragraph{training}
\textit{Anglo-Saxon\_Futhorc,
Armenian,
Atlantean,
Aurek-Besh,
Balinese,
Bengali,
Braille,
Burmese\_(Myanmar),
Cyrillic,
Early\_Aramaic,
Ge\_ez,
Grantha,
Gujarati,
Inuktitut\_(Canadian\_Aboriginal\_Syllabics),
Japanese\_(hiragana),
Japanese\_(katakana),
Kannada,
Keble,
Korean,
Latin,
Malay\_(Jawi\_-\_Arabic),
Malayalam,
Manipuri,
Mkhedruli\_(Georgian),
Ojibwe\_(Canadian\_Aboriginal\_Syllabics),
Sanskrit,
Sylheti,
Syriac\_(Estrangelo),
Tagalog,
Tifinagh
}
\paragraph{validation}
\textit{
Asomtavruli\_(Georgian),
Futurama,
Oriya,
ULOG
}

\paragraph{testing}
\textit{
Alphabet\_of\_the\_Magi,
Angelic,
Arcadian,
Atemayar\_Qelisayer,
Avesta,
Blackfoot\_(Canadian\_Aboriginal\_Syllabics),
Glagolitic,
Greek,
Gurmukhi,
Hebrew,
Mongolian,
N\_Ko,
Old\_Church\_Slavonic\_(Cyrillic),
Syriac\_(Serto),
Tengwar,
Tibetan
}
\section{Tiered-ImageNet}\label{appendix:tiered_supp}
\subsection{Hyperparameters}
Table \ref{tab:tiered_hyperparameters} provides the hyperparameters used for tiered-ImageNet training and testing.

\begin{sidewaystable}
    \centering
    \caption{Tiered-ImageNet hyperparameter summary.}
    \begin{tabular}{l r r r | r r r}
        \toprule
        &\multicolumn{3}{c}{Few-shot} & \multicolumn{3}{c}{Few-task Few-shot} \\
         Hyperparameter & MAML & MALTML  & 
         \begin{tabular}{@{}c@{}}MALTML- \\ Reptile\end{tabular} & MALTML & \begin{tabular}{@{}c@{}}MALTML- \\ ANIL (partial)\end{tabular} & \begin{tabular}{@{}c@{}}MALTML- \\ ANIL (full)\end{tabular} \\
         \midrule
         Model architecture & Conv-4 & Conv-4 & Conv-4 & Conv-4 & Conv-4 & Conv-4\\
         Image input size & $84\times84$ & $84\times84$& $84\times84$& $84\times84$& $84\times84$& $84\times84$\\
         Outer optimizer & Adam & Adam & Adam & Adam & Adam & Adam\\
         Outer step size & 0.001 & 0.001 & 0.001 & 0.001 & 0.001 & 0.001 \\
         Query examples/task & 15 & 15 & 15 & 15 & 15 & 15\\
         Support Tasks/family \textit{(shots)}  & n.a. & 4(1), 2(5) & 4(1), 2(5)  & 4(1), 2(5) & 4(1), 2(5) & 4(1), 2(5)\\
         Query Tasks/family \textit{num(shot)}  & n.a. & 8(1), 4(5) & 8(1), 4(5)  & 8(1), 4(5) & 8(1), 4(5) & 8(1), 4(5)\\
         Training batch size \textit{num(shot)}  & 8(1), 4(5) & 1 family & 1 family & 1 family & 1 family & 1 family\\
         \midrule
         Meta-tuning optimizer  & n.a. & SGD & SGD & SGD & SGD & SGD\\
         Meta-tuning step size  & n.a. & 0.05 & 0.05 & 0.05 & 0.05 & 0.05\\
         Meta-tuning steps  & n.a. & 1 & 1 & 1 & 1 & 1\\
         Meta-tune entire network & \checkmark & \checkmark & \checkmark & \checkmark & $\times$ & \checkmark\\
         \midrule
         Fine-tuning optimizer  & SGD & SGD & SGD & SGD & SGD & SGD\\
         Fine-tuning step size  & 0.5 & 0.5 & 0.5 & 0.5 & 0.5 & 0.5\\
         Fine-tuning steps  & 1 & 1 & 1 & 1 & 1 & 1\\
         Fine-tune entire network & \checkmark & \checkmark & \checkmark & \checkmark & $\times$ & $\times$\\
        \bottomrule
    \end{tabular}
    \label{tab:tiered_hyperparameters}
\end{sidewaystable}

\subsection{training (20), validation (6), test (8) categories}\label{sec:tiered_categories}

\paragraph{training}
\textit{
'game equipment', 'electronic equipment', 'snake, serpent, ophidian', 'tool', 'establishment', 'passerine, passeriform bird', 'aquatic bird', 'primate', 'garment', 'terrier', 'saurian', 'ungulate, hoofed mammal', 'feline, felid', 'restraint, constraint', 'building, edifice', 'musical instrument, instrument', 'instrument', 'protective covering, protective cover, protect', 'hound, hound dog', 'craft'}
\paragraph{validation}
\textit{
'motor vehicle, automotive vehicle', 'furnishing', 'machine', 'durables, durable goods, consumer durables', 'mechanism', 'sporting dog, gun dog'
}

\paragraph{testing}
\textit{
'working dog', 'aquatic vertebrate', 'vessel', 'geological formation, formation', 'obstruction, obstructor, obstructer, impedimen', 'solid', 'substance', 'insect'}
\subsection{Family and task visualization}
\begin{figure}[htbp]
\floatconts
  {fig:tieredfamily}
  {\caption{Visualization of the tiered-ImageNet family distribution}}
  {
    \subfigure[family: `instrument']{
    \addstackgap{\stackunder[5pt]{
    \includegraphics[width=0.2\linewidth]{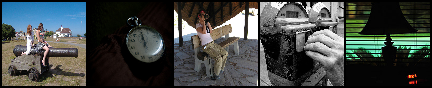}}{task 1}}
    \addstackgap{\stackunder[5pt]{
      \includegraphics[width=0.2\linewidth]{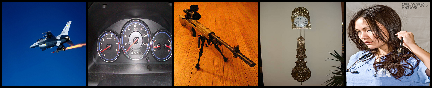}}{task 2}}
}
  \vspace{2mm}
  \qquad
  \subfigure[family: `saurian']{
    \addstackgap{\stackunder[5pt]{
    \includegraphics[width=0.2\linewidth]{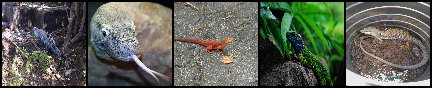}}{task 1}}
    \addstackgap{\stackunder[5pt]{
      \includegraphics[width=0.2\linewidth]{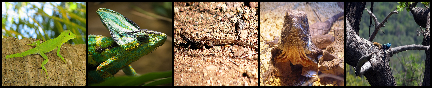}}{task 2}}
  }
  \vspace{2mm}
  \qquad
  \subfigure[family: `musical instrument, instrument']{
    \addstackgap{\stackunder[5pt]{
    \includegraphics[width=0.2\linewidth]{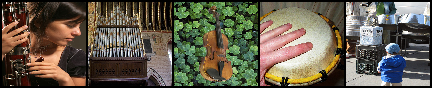}}{task 1}}
   \addstackgap{\stackunder[5pt]{
      \includegraphics[width=0.2\linewidth]{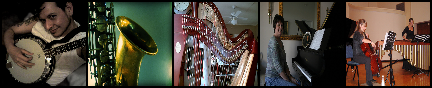}}{task 2}}
  }
  \qquad
  \subfigure[family: 'establishment']{
    \addstackgap{\stackunder[5pt]{
    \includegraphics[width=0.2\linewidth]{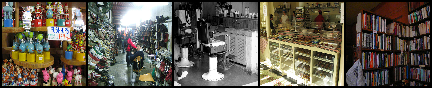}}{task 1}}
   \addstackgap{ \stackunder[5pt]{
      \includegraphics[width=0.2\linewidth]{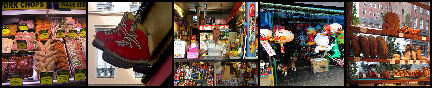}}{task 2}}
  }
  }
\end{figure}

\section{2D navigation}\label{appendix:2d_results}

\begin{figure*}[htbp]
    \subfigure{\label{fig:violin_test_maml}%
      \includegraphics[width=0.32\textwidth]{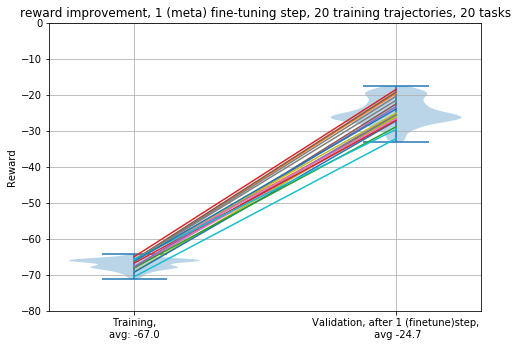}}
    \subfigure{\label{fig:violin_test_trpo}%
        \includegraphics[width=0.32\textwidth]{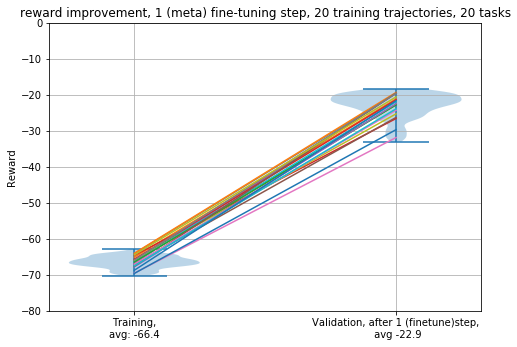}}
    \subfigure{\label{fig:violin_test_metagrad}%
        \includegraphics[width=0.32\textwidth]{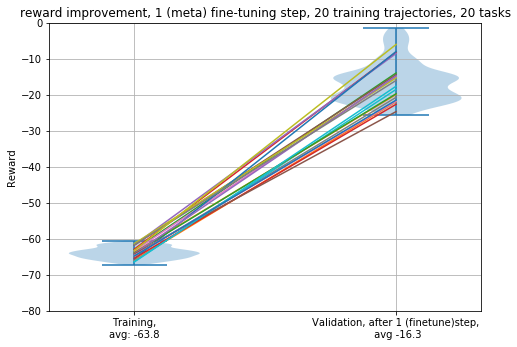}}
\caption{\label{fig:violin} Effect on MAML init of (a) task-finetuning (b) meta-update (TRPO) + task-finetuning (c) meta-update (grad) + task-finetuning.}
\end{figure*}

\begin{figure*}[htbp]
   \subfigure{\label{fig:maml_metagrad_famtuned_metalr}%
      \includegraphics[width=0.49\textwidth]{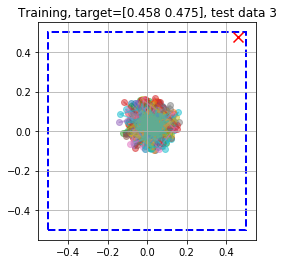}}
   \subfigure{\label{fig:maml_metagrad_tasktuned_metalr0.01}%
      \includegraphics[width=0.49\textwidth]{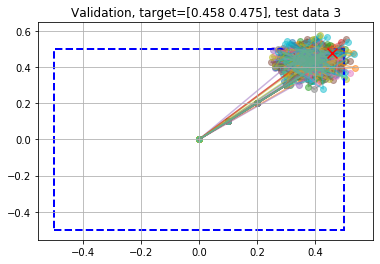}}
\caption{\label{fig:maml_metagrad} Effect on MAML init of (a) 1 gradient meta-tuning step (b) additionally a goal task fine-tuning step. %
}
\end{figure*}
\newpage
\section{Continual Regression}\label{appendix:cont_reg}
\subsection{Hyperparameters}
We borrow the architecture and inner loop hyperparameters from \citet{javed2019metalearning}. For the inner meta-steps, we use a step size of $0.2$. Due to computational constraints, we use only one trajectory each as query and support task and sample one family for each outer update. For each trajectory, we use the same number of tasks (10) and minibatches (40) per task as \citet{javed2019metalearning}. The frequencies, amplitudes, and phases are sampled uniformly from the ranges $[1.0,3.0]$,$[0.1,5.0]$, $[0.0,\pi]$, respectively.
\end{document}